\pdfoutput=1

\documentclass[11pt]{article}

\usepackage[]{emnlp2021}
\usepackage{times}
\usepackage{latexsym}

\usepackage[T1]{fontenc}

\usepackage[utf8]{inputenc}

\usepackage{microtype}

\usepackage{booktabs}
\usepackage{graphicx}
\usepackage{multirow}
\usepackage{amsmath}
\usepackage{subfig}
\usepackage{tabularx}
\usepackage[disable]{todonotes}
\usepackage{wasysym}
\usepackage{enumitem}
\usepackage{float}

\title{TWEAC: Transformer with Extendable QA Agent Classifiers}

\author{Gregor Geigle, Nils Reimers, Andreas R\"uckl\'e, Iryna Gurevych\\ Ubiquitous Knowledge Processing Lab (UKP-TUDA)\\ Department of Computer Science, Technical University of Darmstadt\\ \url{https://www.ukp.tu-darmstadt.de} }

\date{}

\begin{document}
\maketitle
\begin{abstract}

Question answering systems should help users to access knowledge on a broad range of topics and to answer a wide array of different questions. Most systems fall short of this expectation as they are only specialized in one particular setting, e.g., answering factual questions with Wikipedia data. 
To overcome this limitation, we propose composing multiple \emph{QA agents} within a meta-QA system. We argue that there exist a wide range of specialized QA agents in literature. Thus, we address the central research question of how to effectively and efficiently identify suitable QA agents for any given question.
We study both supervised and unsupervised approaches to address this challenge, showing that \emph{TWEAC}---Transformer with Extendable Agent Classifiers---achieves the best performance overall with 94\% accuracy. We provide extensive insights on the scalability of TWEAC, demonstrating that it scales robustly to over 100 QA agents with each providing just 1000 examples of questions they can answer. Our code and data is available.\footnote{\url{https://github.com/UKPLab/TWEAC-qa-agent-selection}}

\end{abstract}

\section{Introduction}
\label{section:intro}

Existing question answering systems are limited in the types of answers they can provide and in the data sources they can access.
For instance, QA systems that rely on a powerful reading comprehension model~\cite[e.g.,][]{yang-etal-2019-end-end-open,chen-etal-2017-reading} cannot answer questions that require access over structured data, e.g., for providing information about the weather (\textit{Will it rain today at 9 am in New York?}) or public transportation (\textit{When is the next train departing from London to Liverpool?}).
Even though recent work has studied how to better generalize QA models across different domains or question types~\cite{ruckle-etal-2020-multicqa,guo2020multireqa,Yang2020MultilingualUS,talmor-berant-2019-multiqa}, 
they investigate only a small fraction of the broad questions that humans may ask (see Table~\ref{tab:exampleq}).

\begin{figure}[t]
    \centering
    \includegraphics[width=0.8\linewidth,]{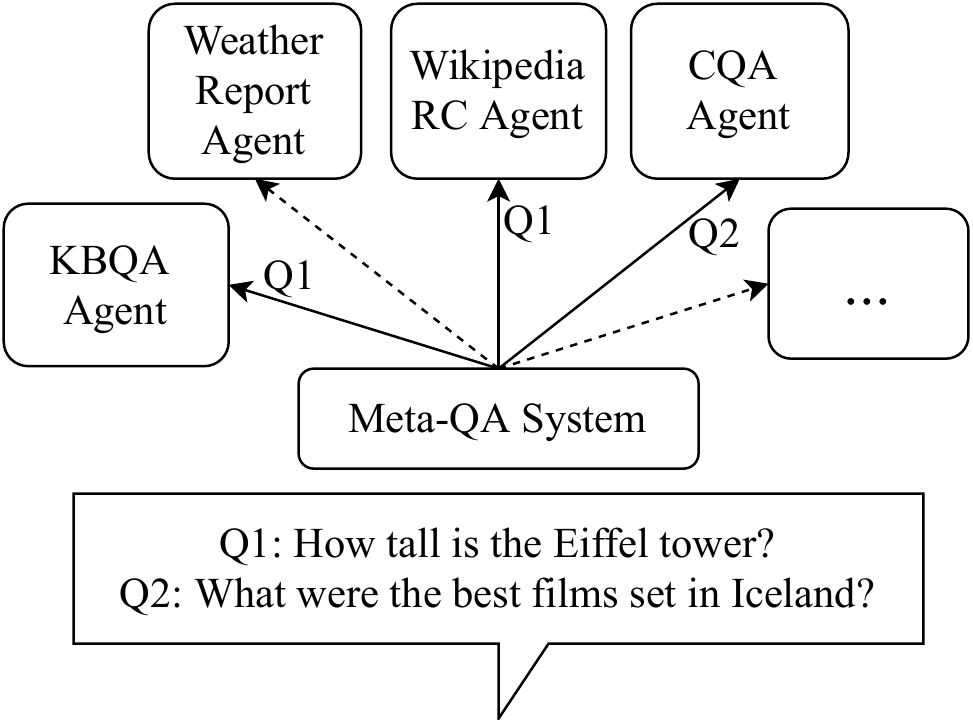}
    \caption{The meta-QA system receives a broad range of questions and needs to decide which QA agent is best suited for answering a question. 
    }
    \label{fig:architecture}
\end{figure}

One approach for answering a much broader range of questions is a \textbf{meta-QA system} that composes various QA agents, i.e., specialized QA subsystems. 
In this work, we address the central research question that arises when developing such a system: how can we efficiently and accurately identify one or multiple QA agents to which we can route the given question (see Figure~\ref{fig:architecture})?

This is conceptually similar to skill selection in dialog systems~\cite{li2019personalized,kim-etal-2018-efficient,burtsev-etal-2018-deeppavlov}, with two central differences: 
(1)~Our setting focuses exclusively on QA and avoids dealing with personalization and dialog history; 
(2)~We are not limited in selecting one skill to address the user's intention, but can invoke multiple QA agents in parallel---e.g., to return multiple answers or to perform consistency checks.

To the best of our knowledge, 
such an approach has not been studied before. 
We, therefore, lay the foundational groundwork for meta-QA systems and focus on the identification of suitable QA agents. 

Importantly, we treat the QA agents as black boxes without further knowledge about their internal workings. This is advantageous because it allows us to study the QA agent selection without external effects from the QA models themselves---e.g., whether they provide a correct answer. We thereby identify the following four challenges for the QA agent selection:
\vspace{2mm}
\begin{enumerate}[nosep,leftmargin=0.5cm]
    \item \textit{Heterogeneous scopes and tasks:} 
    We deal with  
    imbalanced data, various types of questions, etc.  
    \item \textit{Extensibility:} 
    Adding QA agents seamlessly within minutes is crucial to make them available to the meta-QA system as quickly as possible.
    \item \textit{Scalability:} We should be able to handle over one hundred QA agents to cover a broad range of question types, data source, and topics, 
    \item \textit{Small data:} Relevant QA agents should be identified even if they provide only a few examples of questions that they can answer
\end{enumerate}
\vspace{2mm}
To address this, we (1)~create a realistic scenario based on 10 well-known QA tasks, and (2)~construct two substantially larger datasets consisting of questions from 171 subreddits and 200 StackExchange forums. 
In both scenarios, a model receives a question and determines in which data set this question is likely to be observed---thus effectively determining the most appropriate QA agent. While our first scenario is realistic because it covers a wide range of question types that a QA system might receive, our second scenario addresses scalability with over a hundred different agents.

We study two kinds of models that scale well to a large number of QA agents.
(1)~similarity-based models using sentence embeddings and BM25 \citep{robertson2009bm25}, finding that they achieve a high precision without additional fine-tuning;
(2)~Transformer with Extendable Agent Classifiers (TWEAC)---a highly scalable transformer model \cite{vaswani17attention} that is extensible with hundreds of agents without full model re-training. Our results show that TWEAC is the most effective overall and achieves the best performances in both experimental settings. 

We find that our models are very sample efficient---they require only one thousand examples for each QA agent at most to achieve up to 94\% accuracy for identifying the correct one.
Thus, they can be applied to many realistic scenarios in which we do not have large amounts of training data. 
Our models can also scale to many agents and still correctly identify the relevant agent.

We show that our proposed meta-QA system is feasible in realistic and large-scale scenarios even with extending QA agent ensembles and limited data.
This opens up an alternative approach to QA systems whereby we can use multiple QA agents specialized on different questions instead of training a single model to handle all possible questions.

\section{Related Work}

Most of current QA systems are specific to an individual type of QA, e.g., community QA \citep{ruckle-gurevych-2017-end,romeo-etal-2018-flexible}, knowledge-base QA~\citep{sorokin-gurevych-2018-interactive}, or extractive open-domain QA~\citep{yang-etal-2019-end-end-open}.
They implement very narrow QA agents and are not applicable across many different scenarios. 
More recently, research has expanded the scope to multi-domain approaches that can also be applied in zero-shot transfer scenarios~\cite{talmor-berant-2019-multiqa,guo2020multireqa,ruckle-etal-2020-multicqa}. 
Further, some approaches fuse different input sources such as text and knowledge bases~\citep{sun2018,sun-etal-2019-pullnet,xu-etal-2016-hybrid}.
\citet{10.1145/3178876.3186023} propose an adaptable pipeline approach where sub-components in the QA pipeline such as Named Entity Recognition can be selected to best suit the question.
However, none of them address combining fundamentally different QA agents into one extensible system and all are limited to one particular kind of QA such as extracting answer spans from documents.

Consolidating different QA agents into one system is conceptually similar to skill systems in chatbots, where utterances can invoke one of many specialized skills. 
Skill systems are, for instance, part of modern chatbot frameworks, e.g., Deep Pavlov \citep{burtsev-etal-2018-deeppavlov} and ParlAI \citep{miller2017parlai}.
An important research challenge in those scenarios is to better deal with a large and increasing number of skills in combination with user-specific preferences, as it is, for instance, the case in Amazon Alexa~\citep{kim-etal-2018-efficient,Li2019ContinuousLF}.
They also focus on extensibility in their models but their LSTM models require training with hundreds of skills and millions of examples before it can be extended.
Our methods require only hundreds or thousands of available examples.

Identifying QA agents is also related to intent classification \citep{liu2016attention,kato-etal-2017-utterance,gangadharaiah-narayanaswamy-2019-joint,qin-etal-2019-stack} 
and question-type classification \citep{chernov-etal-2015-linguistically,komninos-manandhar-2016-dependency}.
These tasks, however, 
rely on fixed question and intent taxonomies, which limits the extensibility of the developed approaches in practical scenarios.

Identifying QA agents capable of answering a question is related to human expert identification tasks such as CQA Expert Routing \citep{mumaz-et-al-19,li2019personalized} or Reviewer-Paper Matching \citep{zhao2018novel,anjum-etal-2019-pare,DUAN201997}.
They differ from our agent identification as (1) they do not train models for specific experts but instead usually learn a shared embedding space for experts and questions, and (2) human expertise is modeled around the topics the humans are proficient in while our QA agents also differ in the kind of questions asked and not just the topic.

\section{QA Agent Identification}
\label{section:qaagentidentification}
Our goal is to retrieve the QA agents that can---most likely---answer a given question.
In contrast to intent classification~\cite[e.g.,][]{qin-etal-2019-stack} where there is only one correct intent for any given query, we formulate QA agent identification as a ranking task. This is arguably a more realistic approach in our case because one question can potentially be answered by multiple agents.

\subsection{Task Definition}
\label{section:taskdef}

Let $A = \{A_1, A_2, \ldots, A_N\}$ be a set of $N$ different QA agents and let $q$ be a user question. 
Our goal is to rank the agents in $A$ with respect to their (anticipated) ability to answer the question $q$.

For each QA agent $A_i$, 
we obtain a set $E_i = \{e_1, \ldots, e_{n_i}\}$ of example questions that this agent can answer. We use those examples to train the supervised QA agent selection models.

\subsection{Similarity-based Models}

Our central approach is to adapt the k-nearest neighbor algorithm to rank agents in relation to a question $q$.
Given a function $s(q, e)$ that measures the similarity between $q$ and an example $e$ from the example question  set of an agent, we use $s$ to retrieve the top-k most similar examples $e_1, ..., e_k$. Using the examples, we score each agent with

\begin{equation} \label{eq:npm}
    S(A_i) = \frac{1}{|E_i|}\sum_{j=1}^k I_{E_i}(e_j) s(q, e_j)
\end{equation}
where $I_{E_i}$ is the indicator function indicating if the example $e_j$ is in the set of example questions $E_i$.\footnote{$I_{E_i}(e_j)=1$ if $e_j \in E_i$ and 0 otherwise}
 To address different example set sizes, we normalize by dividing with the example set size $|E_i|$ of the respective agent.  

We evaluate two similarity functions for $s$: BM25 \citep{robertson2009bm25} and the dot product between sentence embeddings from the Universal Sentence Encoder QA (USE-QA) model \citep{Yang2020MultilingualUS}, which was trained on a large question-answer dataset. We found empirically that $k=50$ works well regardless of the number of agents.

Extending the similarity-based methods with new agents is efficiently done by adding the example questions from the new agent to the index. However, the approach might lack knowledge about the idiosyncrasies of particular QA domains. Next, we therefore also study a supervised transformer-based approach.

\subsection{Transformer with Extendable Agent Classifiers (TWEAC)}
\begin{figure}
    \centering
    \includegraphics[width=0.7\linewidth]{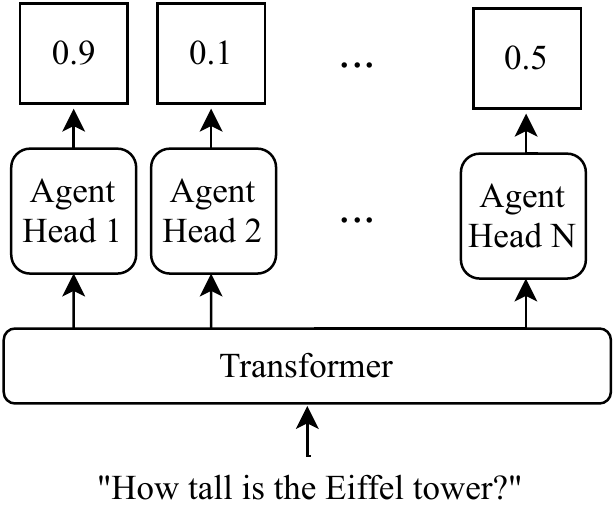}
    \caption{Visualization of TWEAC with $N$ agents}
    \label{fig:model}
\end{figure}

TWEAC consists of a single transformer-based model with a classification head for each agent (see Figure~\ref{fig:model}), which decides whether an agent is capable of answering a question. We experiment with ALBERT \citep{Lan2020ALBERTAL} and RoBERTa \citep{Liu2019RoBERTaAR}.

Each classification head consists of two fully connected layers:
The first layer has a dimensionality of 256 and uses a GELU activation~\citep{Hendrycks2016GaussianEL}.
The second layer outputs a scalar (with sigmoid activation). 
Each head, thus, represents the independent probability of whether one agent can answer the given question. We can then rank the agents by their probabilities. 
We implement the classification heads using two convolution layers with grouped convolutions for parallel execution which gives a sub-linear increase in run-time with respect to the number of classification heads, and it is computationally suitable for classifying thousands of agents.\footnote{
Inference time increases by a factor of 2 when scaling from 10 to 1000 heads and by a factor of 6 when scaling to 5000 heads.}

When adding a new agent to the model, we only introduce a new classification head. 
Afterward, the entire model is fine-tuned with both the training examples of the new agent and examples from previously added agents.
Training with all available data can be infeasible if the model should be ready within minutes after an agent was added. 
We therefore also experiment with a sampling strategy that only uses a fraction of all data in \S\ref{section:extending}. 

We train the model by minimizing the binary cross-entropy (BCE) for each head.
Each training example is considered a positive example for the correct agent and 
as
a negative example for all other agents.
Given the output of the head $h_i$ for agent $A_i$ and the label $y=(y_1, \ldots, y_N)$ as a one-hot encoding of the correct agent, the loss for each head is defined as:
\begin{equation} \label{eq:loss}
    \mathcal{L}_i = -[w_i y_i \log(h_i) + (1-y_i) \log(1-h_i)]
\end{equation}
Weighting of positive examples according to $w_i=(\sum_{j \ne i} |E_j|)/|E_i|$ is necessary to balance the signals from positive and negative examples.

\section{Data and Evaluation}
\label{section:dataeval}

For the evaluation of our models, 
we simulate agents through different datasets without explicitly training a QA agent for the questions. 
This way, we can focus on QA agent identification in an isolated setting without the influence of errors from real QA agents.  
We construct two setups for evaluation: \texttt{QA-Tasks} and \texttt{Many-Agents}.

\subsection{QA Tasks}

\label{section:data_qatasks}
\begin{table}[t]
    \centering
    \footnotesize
\begin{tabularx}{\linewidth}{lX}
\toprule
 \bf QA Agent & \bf  Example Question \\
\midrule
      CQA &  Best browser for web application  \\
      KBQA & Who is the mayor of Tel Aviv \\
      Span RC & Who can enforce European Union law  \\
      Multihop RC & When was the writer of Seesaw born  \\
      Non-factoid QA & How did Beatlemania develop  \\
     Reasoning RC &  Who was first, Edward II or Richard I  \\
     Boolean QA & Is a yard the same as a meter \\
     Claim Validation &  Obama's birth certificate is a forgery\\
     Weather Report &  What will the weather be in Tontitown \\
     Movie Screening & Find the films at Century Theatres  \\
\bottomrule
\end{tabularx}
    \caption{Examples from the 10 QA agents that correspond to 10 common QA tasks in the \texttt{QA-Tasks} dataset.}
    \label{tab:exampleq}
\end{table}

We model a realistic QA scenario with distinct types of questions and minimal overlap between the agents by using questions from different QA datasets from the literature.
We construct the dataset \texttt{QA-Tasks} with 10 agents for different QA tasks.

\textbf{Community QA (CQA)}
is a variant of retrieval-based QA and re-uses the large quantities of questions and answers that have been discussed in question-answering forums~\cite{nakov2017semeval,Tay2017,ruckle-etal-2019-neural,rueckle:AAAI:2019}.
We use question titles from multiple StackExchange forums\footnote{\url{https://archive.org/download/stackexchange}} to cover a range of topics.
StackExchange groups their forums in five categories and we select two forums from each category\footnote{Categories are: Technology, Culture/ Recreation, Life/ Arts, Science, and Professional (\url{https://stackexchange.com/sites}). We include StackOverflow, Superuser, Gaming, English, Stats, Math, Cooking, Photo, Writers, and Workplace.}.
We do not filter the titles so examples can also include just keywords or sentences that are not proper questions. 

\textbf{Knowledge Base QA (KBQA)}
answers factoid questions using a knowledge base \citep{Cui2017KBQALQ}.
We use the questions from QALD-7 \citep{10.1007/978-3-319-69146-6_6} and WebQSP \citep{yih-etal-2016-value} as examples.

\textbf{Span-based Reading Comprehension (Span RC)}
extracts the answer to a question from a corpus of text. 
The examples for this agent come from SQuAD \citep{Rajpurkar2016SQuAD10}

\textbf{Multihop RC}
differs from Span RC as the agent must now perform a multi-step search over multiple documents.
We use HotPotQA \citep{Yang2018HotpotQAAD} as a source for the questions.

\textbf{Non-factoid QA}
requires descriptions or explanations as answers in contrast to factoid questions. We use WikiPassageQA \citep{10.1145/3209978.3210118} as the dataset. 
Different from CQA, which likewise includes non-factoid questions, WikiPassageQA uses Wikipedia data as answers and the questions have been crowd-sourced.

\textbf{Reasoning RC}
covers a range of questions that require higher reasoning like arithmetic or comparisons as proposed by \citet{Dua2019DROPAR}.
We use their DROP dataset for example questions.

\textbf{Boolean QA}
is an RC task with questions that have a yes/no answer as proposed by \citet{clark-etal-2019-boolq}.
We use their BoolQ for the agent. 
The questions are based on Google queries which have a Wikipedia passage that answers the question. 

\textbf{Claim Validation}
fact-checks if a claim is supported by evidence.
We use the Snopes Corpus created by \citet{hanselowski-etal-2019-richly}. 
The difference to Boolean QA is that the claims and the evidence are from Snopes who include potentially unreliable sources including false news sites.

\textbf{Weather Report \& Movie Screening}
are two agents that we use as examples for highly specialized questions which can be seen as skills or intents in conversational agents.
Both agents use examples from the NLU Benchmark by \citet{Coucke2018SnipsVP} for the intents GetWeather and SearchScreeningEvent respectively.

\textbf{Dataset Splits:}
We use the splits of the published datasets. If the test set is unavailable,\footnote{BoolQ, HotPotQA, DROP, SQuAD, NLU Benchmark} we use the development split as the test set and remove the last 25\% of the training data to obtain a new dev set.
We also obtain a dev set for QALD-7 and WebQSP the same way.
For the StackExchange data, each forum is split into train, dev, and test set, such that each split contains all 10 forums.

\subsection{Large-Scale QA Agent Identification}
A meta-QA system as we propose can have a large number of agents, e.g., in a public system where developers can publish their agents.
Some might be general QA agents, while others are highly specialized agents able to answer questions for specific domains or tasks.
We
simulate such a setup using 171 forums from StackExchange and 200 randomly selected subreddits from reddit\footnote{BigQuery (\url{https://console.cloud.google.com/bigquery/}) table fh-bigquery:reddit\_posts}.
We consider all post titles as questions to obtain a sufficient number of examples for each agent.
We term this setup \texttt{Many-Agents}.

It is important to note that this setup is artificial in that we separate agents primarily by topics 
and less by the types of questions.
There is also a higher overlap between questions of different forums:
while each StackExchange forum and subreddit is dedicated to a specific topic, there might be a certain overlap between related topics, e.g., \textit{English} and \textit{English Language Learners} in StackExchange. 
Nevertheless, we argue that this approach allows us to investigate a larger number of agents and how our models can scale to such numbers than would be possible when relying on QA research datasets as in \S\ref{section:data_qatasks}.

\section{Experiments}
\label{section:experiments}

We evaluate the models with respect to their (1)~support of heterogeneous scopes and tasks, (2)~performance with little training data, and (3)~scalability. 
Our two setups from \S\ref{section:dataeval} are created with (1) in mind and our experiments that we introduce in what follows consider
(2) and (3) as well.
Later in \S\ref{section:extending}, we investigate how to \emph{efficiently} extend TWEAC with new agents.
In addition, we provide a comparison of inference speed between our models
for the agent selection in Appendix \S\ref{section:inference}.
\subsection{Experimental Setup}
\label{section:ex_setup1}
\paragraph{Models and Hyperparameters.}
For the similarity-based models, we empirically find that $k=50$ works the best and use this in all experiments. As for similarity functions, we compare BM25 and USE-QA with the dot-product.

For the transformer based ranker (TWEAC), we use \textit{ALBERT base} (denoted as TWEAC{\scriptsize{A-b}}), \textit{ALBERT large} (TWEAC{\scriptsize{A-l}}) and \textit{RoBERTa large} (TWEAC{\scriptsize{R-l}}). 
We train all models for 10 epochs with a batch size of 32, a learning rate of 1e-4 (or 2e-5 for large transformers) with a linear warm-up during the first epoch.
The hyperparameters were chosen after some preliminary training runs on \texttt{QA-Tasks}.

We report Accuracy@1 and mean reciprocal rank (MRR) as performance scores. 
We make the assumption that only the agent from the respective dataset from which we draw the test question is relevant, all other agents are irrelevant. 
This assumption is a result of our dataset construction.

\paragraph{Sample Efficiency.}
We test how many example questions are needed to achieve appropriate ranking results. We evaluate our models on \texttt{QA-Tasks}
with 64 up to 4096 examples per agent.
We note that five agents have less than 4000 examples\footnote{Claim Validation (3847), Non-factoid QA (3331), KBQA (2398), Weather Report \& Movie Screening (226 each)} which 
is accounted for in our weighting component (see Equation~\ref{eq:npm} and \ref{eq:loss}).
TWEAC is trained with all 10 agents at the same time.
We adjust the number of epochs accordingly to keep the number of update steps equal to 10 epochs with 1024 examples.

\paragraph{Scalability.}
We train models in \texttt{Many-Agents} with an increasing number of agents to evaluate the scalability of the different models. For both reddit and StackExchange, we train a model with 10, 50, 100 and 171 (StackExchange)/ 200 (reddit) agents.
TWEAC is newly trained for each set of agents (using 1024 examples per agent).

\subsection{Results}

\paragraph{Sample Efficiency.}
\begin{figure}
    \centering
    \includegraphics[width=\linewidth,height=4.8cm]{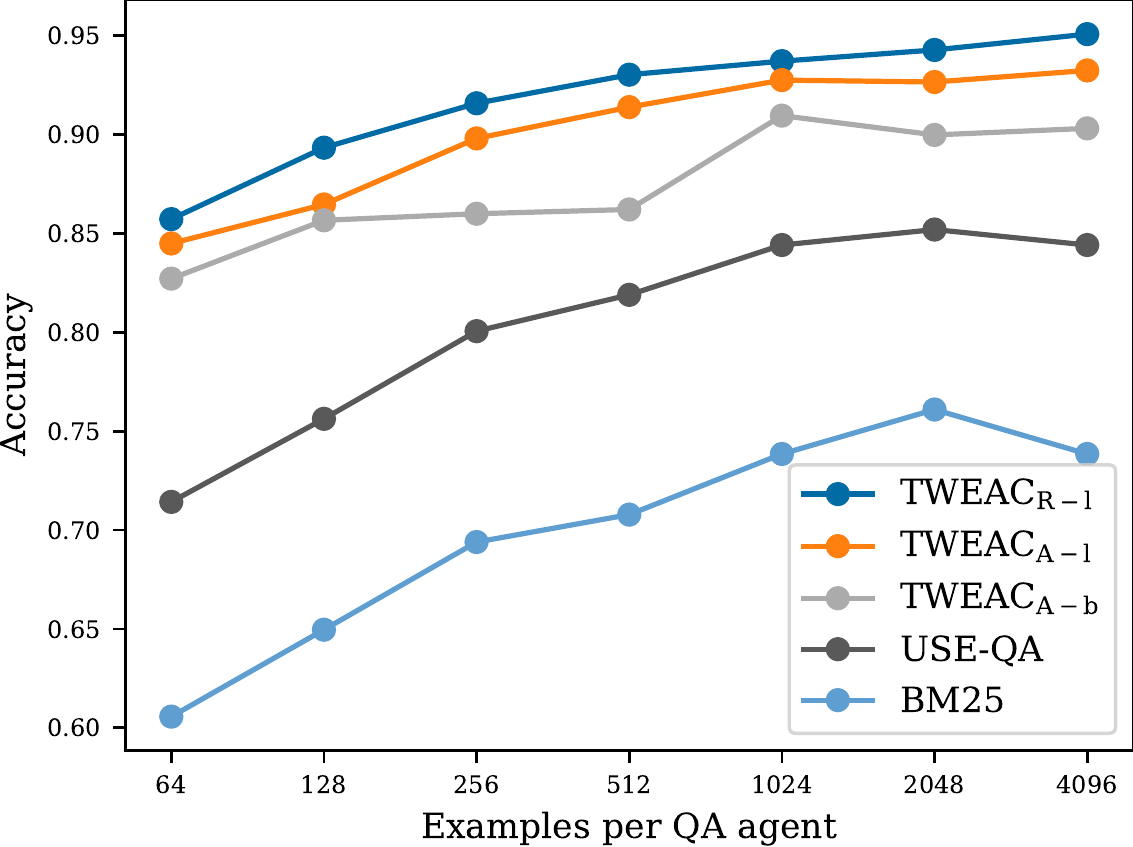}
    \caption{Accuracy of our models trained on \texttt{QA-Tasks} with increasing number of training examples per agent.}
    \label{fig:sample}
\end{figure}

Figure~\ref{fig:sample} shows the performance of methods in \texttt{QA-Tasks} as a function of the number of training examples per agent.\footnote{Precise numbers for these figures can be found in the respective tables in the appendix.} 

The TWEAC models achieve with 64 examples already over 80\% accuracy and over 90\% with a thousand examples.
Similarity-based models perform noticeably worse---especially with very few examples---but reduce the gap to TWEAC with a larger number of examples.
Starting at 1024, more examples have a diminishing return for TWEAC and might negatively impact the similarity-based models. Independent of how large we select the example sets for the agents, we observe the same performance ranks for the five evaluated systems  with TWEAC{\scriptsize{R-l}} performing best and BM25 performing worst.

Inspecting the performances of individual agents, we find that highly specialized agents such as Weather Report and Movie Screening perform very well ($\geq$95\% accuracy) with very few examples.
QA Agents with a wider range of possible questions or topics like CQA or Span RC, on the other hand, are disproportionately worse with few examples.
However, the need for more examples with such agents is often no problem in practice as they likely require more 
training data for the agent subsystem and, thus, have more examples available.

\paragraph{Scalability.}
\label{section:scale}
\begin{figure}[t]
    \centering
    \subfloat[reddit]{\includegraphics[width=\linewidth,height=4.8cm]{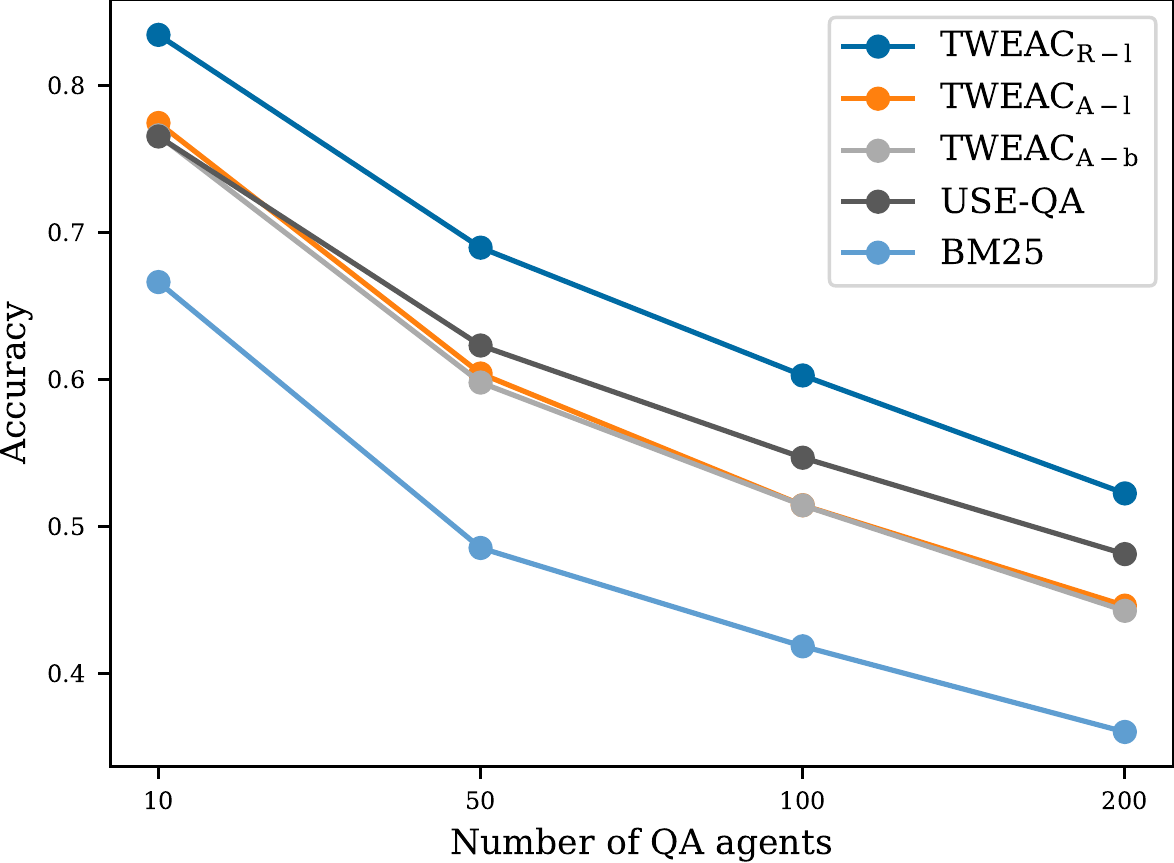}} \\
    \subfloat[StackExchange]{\includegraphics[width=\linewidth,height=4.8cm]{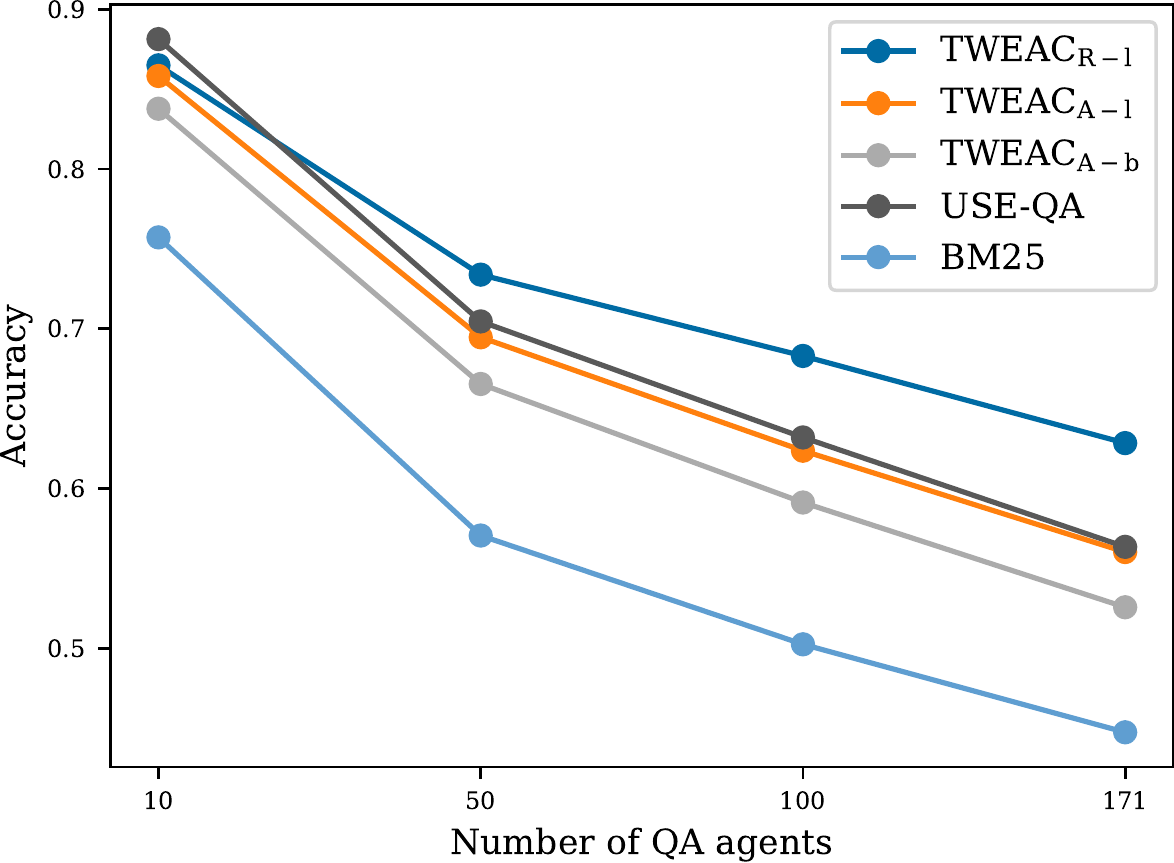}}
    \caption{Accuracy of models trained with an increasing number of agents from \texttt{Many-Agents}}
    \label{fig:setup2comp}
\end{figure}
We present the results of the large-scale experiments with \texttt{Many-Agents} in Figure~\ref{fig:setup2comp} as a function of the number of agents.\footnote{Detailed results are given in the appendix.}
In contrast to the previous experiment, we now observe rank changes between our methods when increasing the number of agents. 
In particular, USE-QA achieves better results compared to TWEAC{\scriptsize{A-b}} and TWEAC{\scriptsize{A-l}}, and with a smaller number of 10 agents it also outperforms  TWEAC{\scriptsize{R-l}} on the StackExchange dataset.
A potential reason for the better performance of USE-QA with \texttt{Many-Agents} as compared to \texttt{QA-Tasks} is that USE-QA has been trained on reddit and StackExchange data.

We observe similar performance drops across all models ranging from 20 to 30 points accuracy and MRR when increasing the agents from 10 to the maximum number. This is expected, as the task difficulty increases with a larger pool of agents. Further, as we analyze in Section \ref{section:analysis}, most mistakes stem from similar topics, e.g., the agent for the Math StackExchange ranked higher than the Statistics StackExchange agent for a statistics question. Even for a large pool of agents, we still achieve high Accuracy@1 results compared to a random baseline, and the correct agent is ranked on average at least second with all models (except BM25).

\paragraph{Summary.}

Our proposed models manage to identify the correct agents with high precision in a realistic setup with 10 QA tasks while also scaling well to hundreds of agents.
They can achieve good performances even with as little as 64 training examples per agent.
TWEAC---especially with large transformers---outperforms similarity-based models in most of our experiments.

\section{Extending TWEAC}
\label{section:extending}
In a realistic setting, we may want to add new QA agents over time.
Similarity-based models only need to index the new examples.
Extending TWEAC 
is comparatively harder as it needs to be fine-tuned for the new agent. Re-training on all available data each time we add a new agent would 
be prohibitive
and it would take hours until a new agent is eventually integrated into the system.

We consider a \textit{half-and-half} sampling strategy that uses a constant number of examples regardless of the number of agents.
It uses all 1024 examples from the extended agent but randomly selects $1024/(|\text{agents}|-1)$ examples from each of the other agents where $|\text{agents}|$ is the number of agents in the model for a total of approximately 2048 examples in total.
Each epoch, a new set of examples is selected.
We hypothesize that just a few examples from the previously added agents are necessary as (1)~the heads were already trained with the examples in previous iterations and (2)~the new head requires just a subset of the negative examples to distinguish them from its own examples.

The baseline \textit{no sampling} uses all available examples for each agent 
when extending the model.
With \textit{full training}, we train a new model from scratch with all agents and do not extend a previously trained model.

\paragraph{Extending TWEAC Once.}

We extend the model from 9 to 10 agents in the \texttt{QA-Tasks} setup.
We evaluate this with a leave-one-out approach:
We train TWEAC with 9 of the agents and extend with the remaining one.
We repeat this for each agent and report the average results.
During extension, we train the entire model including the transformer and all classification heads.
We compare against a model that was fine-tuned with examples from all 10 agents.  
We present the results in Table \ref{tab:extension}.

\begin{table}[t]
\centering
\footnotesize
\begin{tabular}{lrr}
\toprule
         \bf Strategy &  \bf Accuracy & \bf MRR \\
          \midrule
Full training    &    0.9096 & 0.9462 \\
No sampling &   0.9105 & 0.9467 \\
Half-and-half & 0.8860 & 0.9311 \\
\bottomrule
\end{tabular}
    \caption{Average performance of extending a model from 9 to 10 agents from \texttt{QA-Tasks} with no sampling or half-and-half sampling compared to a model directly trained on all 10 agents (full training).}
    \label{tab:extension}
\end{table}

We see that extending with \textit{no sampling} performs as well as the model trained directly with all 10 agents.
The model trained with \textit{half-and-half sampling} slightly drops in accuracy by 2 points.

\paragraph{Iterative Extension.}
We iteratively extending TWEAK one agent at a time with the \texttt{Many-Agents} datasets.
We start with 10 or 50 agents and extend to 100.

\begin{figure}[t]
    \centering
    \subfloat[reddit]{\includegraphics[width=\linewidth,height=4.8cm]{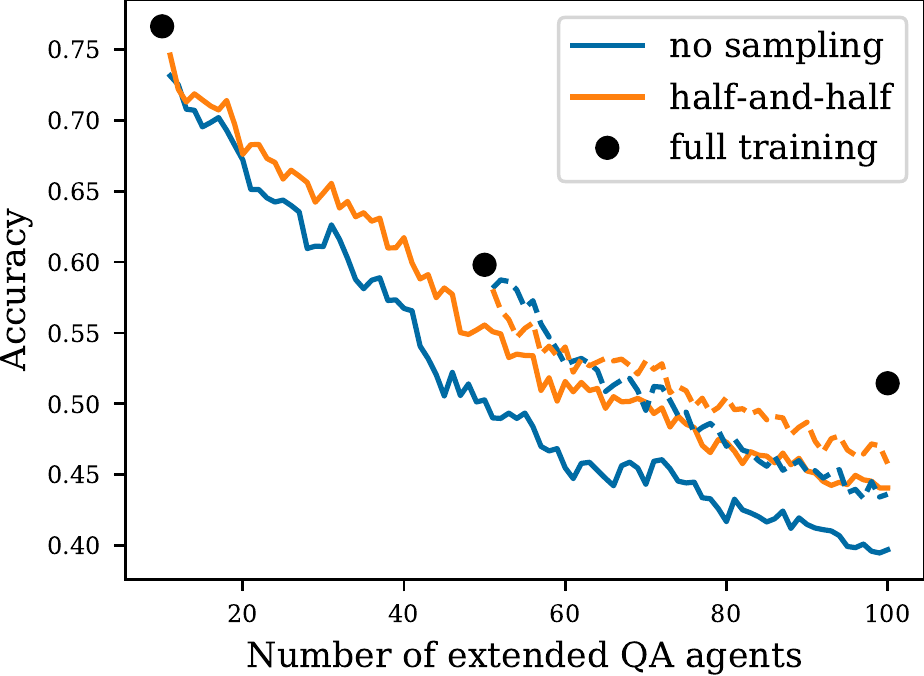}} \\
    \subfloat[StackExchange]{\includegraphics[width=\linewidth,height=4.8cm]{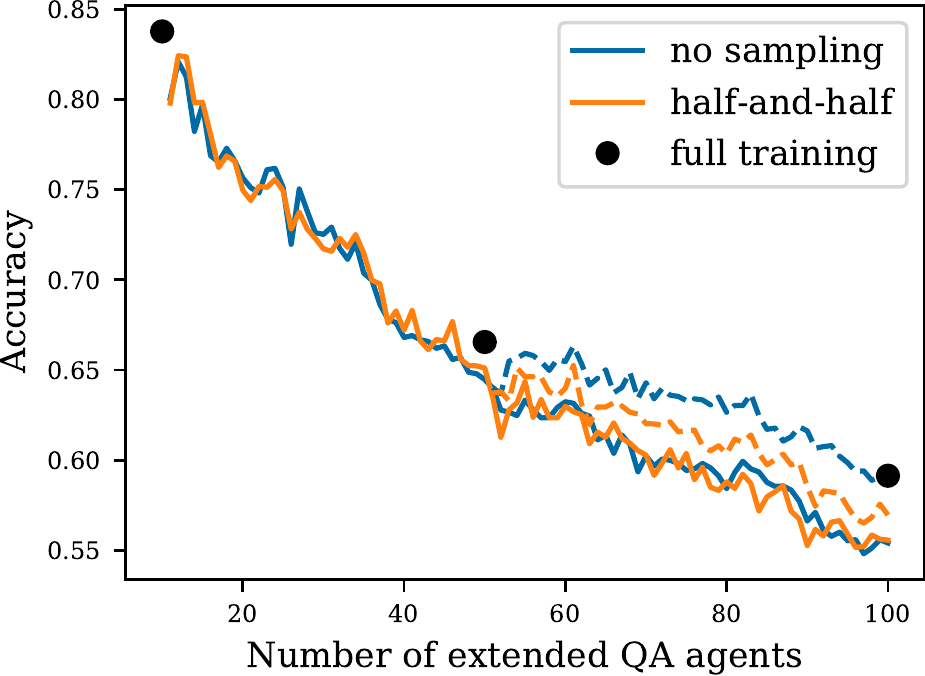}}
    \caption{Accuracy of Iterative Extension with \texttt{Many-Agents}. Solid line starts
    with 10 agents and dashed line starts with 50 agents. Dots show performance of \textit{full training}
    .}
    \label{fig:it_ext}
\end{figure}

We plot the accuracy as a function of the number of extended agents in Figure~\ref{fig:it_ext}.\footnote{Exact numbers can be found in the tables in the appendix.}
After iterative extension from 10 to 100 agents, \textit{half-and-half sampling} is 4 points better for reddit and on-par with StackExchange compared to \textit{no sampling}.
The gap between the two with the model starting at 50 agents is 2 points for both datasets.
\textit{Half-and-half sampling} can thus be used to train models with only a fraction of all examples without a decrease in performance compared to using all examples.
However, models trained directly with all agents attain higher scores than iterative extension with \textit{half-and-half sampling}.
The gap between a model trained with all 100 agents and \textit{half-and-half} starting at 10 agents is 7/ 4 points but it shrinks to 5/ 2 points when starting with 50 agents.

\paragraph{Summary.}
Out of the analyzed options, \textit{full training} TWEAC from scratch yields the highest accuracy. However, the training time increases linearly with the number of QA agents. 
Using \textit{half-and-half sampling}, we can rapidly extend the model with new QA agents. It has roughly a constant training time independent of the number of previously existent QA agents. The resulting slight drop in performance can be reduced by periodically 
\textit{full training} a model with all agents.

\section{Analysis}
\label{section:analysis}

The scalability experiments in \S\ref{section:scale} showed a performance decrease of all models with more agents.
We analyze the errors made by TWEAC{\scriptsize{R-l}} trained on 200 agents from reddit and  171 agents from StackExchange to assess what causes this decrease.
We focus on the misclassification errors at rank 1 and count the errors for the different unique mistakes symmetrical,
 i.e., errors that mistake agent A for agent B and the other way around are both counted towards A-B mistakes.

The majority of errors are due to only a few mistakes. 
In both datasets, around 0.5\% of all  unique mistakes account for 10\% of all errors and every second error is due to less than 10\% of all mistakes (detailed numbers in the appendix).
The mistakes with the largest number of errors are for StackExchange \textit{english}-\textit{ell} (english language learner), \textit{cstheory}-\textit{cs} and \textit{space}-\textit{astronomy}. For reddit it is \textit{politics}-\textit{Conservative}, \textit{recipes}-\textit{food} and \textit{Autos}-\textit{cars}.

These mistakes are due to selecting agents of forums with very similar topics, e.g., \textit{english} and \textit{ell} overlap considerably in the topics users address in these forums. 
Mistaking such agents is a reasonable error to make and we show that humans also fail to correctly identify the correct agent for questions where the model failed:
We randomly select 50 errors made by the model each from reddit and StackExchange and ask three expert annotators to select the right agent between the true label and the (wrong) predicted label.
On average, they achieve 43\% accuracy for reddit and 55\% accuracy for StackExchange, indicating that humans are not better than chance to identify the right forum for these questions.

In summary, the lower accuracy with a large number of agents is mainly caused by overlapping agents.
The evaluation metrics cannot take the overlapping agents into account as we are limited to only one right agent.
It is therefore important for future work to design better datasets that take into account that multiple agents can answer a question.

\section{Conclusion}
We analyzed how to automatically select suitable QA agents specializing in different questions, as an alternative to a single QA system that tries to cover all possible questions.
We presented a scalable meta-QA system that allows for a flexible extension with different QA agents. For newly posed questions, we rank agents by their ability to answer them and select the most suitable ones.

To evaluate the QA agent selection task, we created a realistic scenario with QA agents based on 
different QA tasks, and constructed a large-scale setting with hundreds of agents using public forums from reddit and StackExchange.
We have established that similarity-based methods and our newly proposed approach TWEAC are scalable, extensible, and sample efficient. 
To allow for a fast integration of new QA agents, we presented a \textit{half-and-half} sampling strategy, which extends TWEAC without full re-training using just a fraction of all available training data.

Future work could further explore the overlap between QA agents with respect to the questions they can answer, e.g.,  by creating dedicated datasets on QA agent selection. Our current datasets do not account for such overlap, potentially 
under-estimating the models' capabilities. 

\section*{Acknowledgments}

This work has been support by multiple sources. (1) The German Research Foundation through the German-Israeli Project Cooperation (DIP, grant DA 1600/1-1 and grant GU 798/17-1),  (2) by the German Federal Ministry of Education and Research and the Hessian Ministry of Higher Education, Research, Science and the Arts within their joint support of the National Research Center for Applied Cybersecurity ATHENE, (3) by European Regional Development Fund (ERDF) and the Hessian State Chancellery – Hessian Minister of Digital Strategy and Development under the promotional reference 20005482 (TexPrax), (4) the German Research
Foundation (DFG) as part of the UKP-SQuARE project (grant GU 798/29-1).

We thank Jonas Pfeiffer, Tilman Beck and Leonardo Ribeiro for their insightful feedback and suggestions on a draft of this paper.

\bibliography{emnlp2021}

\begin{thebibliography}{47}
\expandafter\ifx\csname natexlab\endcsname\relax\def\natexlab#1{#1}\fi

\bibitem[{Anjum et~al.(2019)Anjum, Gong, Bhat, Hwu, and
  Xiong}]{anjum-etal-2019-pare}
Omer Anjum, Hongyu Gong, Suma Bhat, Wen-Mei Hwu, and JinJun Xiong. 2019.
\newblock \href {https://doi.org/10.18653/v1/D19-1049} {{P}a{R}e: A
  paper-reviewer matching approach using a common topic space}.
\newblock In \emph{Proceedings of the 2019 Conference on Empirical Methods in
  Natural Language Processing and the 9th International Joint Conference on
  Natural Language Processing (EMNLP-IJCNLP)}, pages 518--528, Hong Kong,
  China. Association for Computational Linguistics.

\bibitem[{Burtsev et~al.(2018)Burtsev, Seliverstov, Airapetyan, Arkhipov,
  Baymurzina, Bushkov, Gureenkova, Khakhulin, Kuratov, Kuznetsov, Litinsky,
  Logacheva, Lymar, Malykh, Petrov, Polulyakh, Pugachev, Sorokin, Vikhreva, and
  Zaynutdinov}]{burtsev-etal-2018-deeppavlov}
Mikhail~S. Burtsev, Alexander~V. Seliverstov, Rafael Airapetyan, Mikhail
  Arkhipov, Dilyara Baymurzina, Nickolay Bushkov, Olga Gureenkova, Taras
  Khakhulin, Yuri Kuratov, Denis Kuznetsov, Alexey Litinsky, Varvara Logacheva,
  Alexey Lymar, Valentin Malykh, Maxim Petrov, Vadim Polulyakh, Leonid
  Pugachev, Alexey Sorokin, Maria Vikhreva, and Marat Zaynutdinov. 2018.
\newblock \href {https://doi.org/10.18653/v1/P18-4021} {Deep{P}avlov:
  Open-source library for dialogue systems}.
\newblock In \emph{Proceedings of {ACL} 2018, Melbourne, Australia, July 15-20,
  2018, System Demonstrations}, pages 122--127. Association for Computational
  Linguistics.

\bibitem[{Chen et~al.(2017)Chen, Fisch, Weston, and
  Bordes}]{chen-etal-2017-reading}
Danqi Chen, Adam Fisch, Jason Weston, and Antoine Bordes. 2017.
\newblock \href {https://doi.org/10.18653/v1/P17-1171} {Reading {W}ikipedia to
  answer open-domain questions}.
\newblock In \emph{Proceedings of the 55th Annual Meeting of the Association
  for Computational Linguistics (Volume 1: Long Papers)}, pages 1870--1879,
  Vancouver, Canada. Association for Computational Linguistics.

\bibitem[{Chernov et~al.(2015)Chernov, Petukhova, and
  Klakow}]{chernov-etal-2015-linguistically}
Alexandr Chernov, Volha Petukhova, and Dietrich Klakow. 2015.
\newblock \href
  {http://www.ep.liu.se/ecp/article.asp?issue=109\&article=009\&volume=}
  {Linguistically motivated question classification}.
\newblock In \emph{Proceedings of the 20th Nordic Conference of Computational
  Linguistics, {NODALIDA} 2015, Institute of the Lithuanian Language, Vilnius,
  Lithuania, May 11-13, 2015}, volume 109 of \emph{Link{\"{o}}ping Electronic
  Conference Proceedings}, pages 51--59. Link{\"{o}}ping University Electronic
  Press / Association for Computational Linguistics.

\bibitem[{Clark et~al.(2019)Clark, Lee, Chang, Kwiatkowski, Collins, and
  Toutanova}]{clark-etal-2019-boolq}
Christopher Clark, Kenton Lee, Ming-Wei Chang, Tom Kwiatkowski, Michael
  Collins, and Kristina Toutanova. 2019.
\newblock \href {https://doi.org/10.18653/v1/N19-1300} {{B}ool{Q}: Exploring
  the surprising difficulty of natural yes/no questions}.
\newblock In \emph{Proceedings of the 2019 Conference of the North {A}merican
  Chapter of the Association for Computational Linguistics: Human Language
  Technologies, Volume 1 (Long and Short Papers)}, pages 2924--2936,
  Minneapolis, Minnesota. Association for Computational Linguistics.

\bibitem[{Cohen et~al.(2018)Cohen, Yang, and Croft}]{10.1145/3209978.3210118}
Daniel Cohen, Liu Yang, and W.~Bruce Croft. 2018.
\newblock \href {https://doi.org/10.1145/3209978.3210118} {Wikipassageqa: A
  benchmark collection for research on non-factoid answer passage retrieval}.
\newblock In \emph{The 41st International ACM SIGIR Conference on Research \&
  Development in Information Retrieval}, SIGIR '18, page 1165–1168, New York,
  NY, USA. Association for Computing Machinery.

\bibitem[{Coucke et~al.(2018)Coucke, Saade, Ball, Bluche, Caulier, Leroy,
  Doumouro, Gisselbrecht, Caltagirone, Lavril, Primet, and
  Dureau}]{Coucke2018SnipsVP}
Alice Coucke, Alaa Saade, Adrien Ball, Th{\'{e}}odore Bluche, Alexandre
  Caulier, David Leroy, Cl{\'{e}}ment Doumouro, Thibault Gisselbrecht,
  Francesco Caltagirone, Thibaut Lavril, Ma{\"{e}}l Primet, and Joseph Dureau.
  2018.
\newblock \href {http://arxiv.org/abs/1805.10190} {Snips voice platform: an
  embedded spoken language understanding system for private-by-design voice
  interfaces}.
\newblock \emph{arXiv preprint}, abs/1805.10190.

\bibitem[{Cui et~al.(2017)Cui, Xiao, Wang, Song, Hwang, and
  Wang}]{Cui2017KBQALQ}
Wanyun Cui, Yanghua Xiao, Haixun Wang, Yangqiu Song, Seung{-}won Hwang, and Wei
  Wang. 2017.
\newblock \href {https://doi.org/10.14778/3055540.3055549} {{KBQA:} learning
  question answering over {QA} corpora and knowledge bases}.
\newblock \emph{Proceedings of the {VLDB} Endowment}, 10(5):565--576.

\bibitem[{Dua et~al.(2019)Dua, Wang, Dasigi, Stanovsky, Singh, and
  Gardner}]{Dua2019DROPAR}
Dheeru Dua, Yizhong Wang, Pradeep Dasigi, Gabriel Stanovsky, Sameer Singh, and
  Matt Gardner. 2019.
\newblock \href {https://doi.org/10.18653/v1/N19-1246} {{DROP}: A reading
  comprehension benchmark requiring discrete reasoning over paragraphs}.
\newblock In \emph{Proceedings of the 2019 Conference of the North {A}merican
  Chapter of the Association for Computational Linguistics: Human Language
  Technologies, Volume 1 (Long and Short Papers)}, pages 2368--2378,
  Minneapolis, Minnesota. Association for Computational Linguistics.

\bibitem[{Duan et~al.(2019)Duan, Tan, Zhao, Wang, Chen, and Zhang}]{DUAN201997}
Zhen Duan, Shicheng Tan, Shu Zhao, Qianqian Wang, Jie Chen, and Yanping Zhang.
  2019.
\newblock \href {https://doi.org/https://doi.org/10.1016/j.neucom.2019.06.074}
  {Reviewer assignment based on sentence pair modeling}.
\newblock \emph{Neurocomputing}, 366:97 -- 108.

\bibitem[{Gangadharaiah and
  Narayanaswamy(2019)}]{gangadharaiah-narayanaswamy-2019-joint}
Rashmi Gangadharaiah and Balakrishnan Narayanaswamy. 2019.
\newblock \href {https://doi.org/10.18653/v1/N19-1055} {Joint multiple intent
  detection and slot labeling for goal-oriented dialog}.
\newblock In \emph{Proceedings of the 2019 Conference of the North {A}merican
  Chapter of the Association for Computational Linguistics: Human Language
  Technologies, Volume 1 (Long and Short Papers)}, pages 564--569, Minneapolis,
  Minnesota. Association for Computational Linguistics.

\bibitem[{Guo et~al.(2020)Guo, Yang, Cer, Shen, and
  Constant}]{guo2020multireqa}
Mandy Guo, Yinfei Yang, Daniel Cer, Qinlan Shen, and Noah Constant. 2020.
\newblock \href {http://arxiv.org/abs/2005.02507} {Multireqa: {A} cross-domain
  evaluation for retrieval question answering models}.
\newblock \emph{arXiv preprint}, abs/2005.02507.

\bibitem[{Hanselowski et~al.(2019)Hanselowski, Stab, Schulz, Li, and
  Gurevych}]{hanselowski-etal-2019-richly}
Andreas Hanselowski, Christian Stab, Claudia Schulz, Zile Li, and Iryna
  Gurevych. 2019.
\newblock \href {https://doi.org/10.18653/v1/K19-1046} {A richly annotated
  corpus for different tasks in automated fact-checking}.
\newblock In \emph{Proceedings of the 23rd Conference on Computational Natural
  Language Learning (CoNLL)}, pages 493--503, Hong Kong, China. Association for
  Computational Linguistics.

\bibitem[{Hendrycks and Gimpel(2016)}]{Hendrycks2016GaussianEL}
Dan Hendrycks and Kevin Gimpel. 2016.
\newblock \href {http://arxiv.org/abs/1606.08415} {Bridging nonlinearities and
  stochastic regularizers with gaussian error linear units}.
\newblock \emph{arXiv preprint}, abs/1606.08415.

\bibitem[{Kato et~al.(2017)Kato, Nagai, Noda, Sumitomo, Wu, and
  Yamamoto}]{kato-etal-2017-utterance}
Tsuneo Kato, Atsushi Nagai, Naoki Noda, Ryosuke Sumitomo, Jianming Wu, and
  Seiichi Yamamoto. 2017.
\newblock \href {https://doi.org/10.18653/v1/w17-5508} {Utterance intent
  classification of a spoken dialogue system with efficiently untied recursive
  autoencoders}.
\newblock In \emph{Proceedings of the 18th Annual SIGdial Meeting on Discourse
  and Dialogue, Saarbr{\"{u}}cken, Germany, August 15-17, 2017}, pages 60--64.
  Association for Computational Linguistics.

\bibitem[{Kim et~al.(2018)Kim, Kim, Kumar, and
  Sarikaya}]{kim-etal-2018-efficient}
Young{-}Bum Kim, Dongchan Kim, Anjishnu Kumar, and Ruhi Sarikaya. 2018.
\newblock \href {https://doi.org/10.18653/v1/P18-1206} {Efficient large-scale
  neural domain classification with personalized attention}.
\newblock In \emph{Proceedings of the 56th Annual Meeting of the Association
  for Computational Linguistics, {ACL} 2018, Melbourne, Australia, July 15-20,
  2018, Volume 1: Long Papers}, pages 2214--2224. Association for Computational
  Linguistics.

\bibitem[{Komninos and Manandhar(2016)}]{komninos-manandhar-2016-dependency}
Alexandros Komninos and Suresh Manandhar. 2016.
\newblock \href {https://doi.org/10.18653/v1/N16-1175} {Dependency based
  embeddings for sentence classification tasks}.
\newblock In \emph{Proceedings of the 2016 Conference of the North {A}merican
  Chapter of the Association for Computational Linguistics: Human Language
  Technologies}, pages 1490--1500, San Diego, California. Association for
  Computational Linguistics.

\bibitem[{Lan et~al.(2020)Lan, Chen, Goodman, Gimpel, Sharma, and
  Soricut}]{Lan2020ALBERTAL}
Zhenzhong Lan, Mingda Chen, Sebastian Goodman, Kevin Gimpel, Piyush Sharma, and
  Radu Soricut. 2020.
\newblock \href {https://openreview.net/forum?id=H1eA7AEtvS} {{ALBERT:} {A}
  lite {BERT} for self-supervised learning of language representations}.
\newblock In \emph{8th International Conference on Learning Representations,
  {ICLR} 2020, Addis Ababa, Ethiopia, April 26-30, 2020}. OpenReview.net.

\bibitem[{Li et~al.(2019{\natexlab{a}})Li, Lee, Mudgal, Sarikaya, and
  Kim}]{Li2019ContinuousLF}
Han Li, Jihwan Lee, Sidharth Mudgal, Ruhi Sarikaya, and Young-Bum Kim.
  2019{\natexlab{a}}.
\newblock \href {https://doi.org/10.18653/v1/N19-1379} {Continuous learning for
  large-scale personalized domain classification}.
\newblock pages 3784--3794.

\bibitem[{Li et~al.(2019{\natexlab{b}})Li, Jiang, Sun, and
  Wang}]{li2019personalized}
Zeyu Li, Jyun{-}Yu Jiang, Yizhou Sun, and Wei Wang. 2019{\natexlab{b}}.
\newblock \href {https://doi.org/10.1609/aaai.v33i01.3301192} {Personalized
  question routing via heterogeneous network embedding}.
\newblock In \emph{The Thirty-Third {AAAI} Conference on Artificial
  Intelligence, {AAAI} 2019, The Thirty-First Innovative Applications of
  Artificial Intelligence Conference, {IAAI} 2019, The Ninth {AAAI} Symposium
  on Educational Advances in Artificial Intelligence, {EAAI} 2019, Honolulu,
  Hawaii, USA, January 27 - February 1, 2019}, pages 192--199. {AAAI} Press.

\bibitem[{Liu and Lane(2016)}]{liu2016attention}
Bing Liu and Ian Lane. 2016.
\newblock \href {https://doi.org/10.21437/Interspeech.2016-1352}
  {Attention-based recurrent neural network models for joint intent detection
  and slot filling}.
\newblock In \emph{Interspeech 2016, 17th Annual Conference of the
  International Speech Communication Association, San Francisco, CA, USA,
  September 8-12, 2016}, pages 685--689. {ISCA}.

\bibitem[{Liu et~al.(2019)Liu, Ott, Goyal, Du, Joshi, Chen, Levy, Lewis,
  Zettlemoyer, and Stoyanov}]{Liu2019RoBERTaAR}
Yinhan Liu, Myle Ott, Naman Goyal, Jingfei Du, Mandar Joshi, Danqi Chen, Omer
  Levy, Mike Lewis, Luke Zettlemoyer, and Veselin Stoyanov. 2019.
\newblock \href {http://arxiv.org/abs/1907.11692} {Roberta: {A} robustly
  optimized {BERT} pretraining approach}.
\newblock \emph{arXiv preprint}, abs/1907.11692.

\bibitem[{Miller et~al.(2017)Miller, Feng, Batra, Bordes, Fisch, Lu, Parikh,
  and Weston}]{miller2017parlai}
Alexander~H. Miller, Will Feng, Dhruv Batra, Antoine Bordes, Adam Fisch, Jiasen
  Lu, Devi Parikh, and Jason Weston. 2017.
\newblock \href {https://doi.org/10.18653/v1/d17-2014} {Parlai: {A} dialog
  research software platform}.
\newblock In \emph{Proceedings of the 2017 Conference on Empirical Methods in
  Natural Language Processing, {EMNLP} 2017, Copenhagen, Denmark, September
  9-11, 2017 - System Demonstrations}, pages 79--84. Association for
  Computational Linguistics.

\bibitem[{Mumtaz et~al.(2019)Mumtaz, Rodr{\'{\i}}guez, and
  Benatallah}]{mumaz-et-al-19}
Sara Mumtaz, Carlos Rodr{\'{\i}}guez, and Boualem Benatallah. 2019.
\newblock \href {https://doi.org/10.1007/978-3-030-21290-2\_14} {Expert2vec:
  Experts representation in community question answering for question routing}.
\newblock In \emph{Advanced Information Systems Engineering - 31st
  International Conference, CAiSE 2019, Rome, Italy, June 3-7, 2019,
  Proceedings}, volume 11483 of \emph{Lecture Notes in Computer Science}, pages
  213--229. Springer.

\bibitem[{Nakov et~al.(2017)Nakov, Hoogeveen, M{\`a}rquez, Moschitti, Mubarak,
  Baldwin, and Verspoor}]{nakov2017semeval}
Preslav Nakov, Doris Hoogeveen, Llu{\'\i}s M{\`a}rquez, Alessandro Moschitti,
  Hamdy Mubarak, Timothy Baldwin, and Karin Verspoor. 2017.
\newblock \href {https://www.aclweb.org/anthology/S17-2003} {Semeval-2017 task
  3: Community question answering}.
\newblock In \emph{Proceedings of the 11th International Workshop on Semantic
  Evaluation (SemEval-2017)}, pages 27--48.

\bibitem[{Qin et~al.(2019)Qin, Che, Li, Wen, and Liu}]{qin-etal-2019-stack}
Libo Qin, Wanxiang Che, Yangming Li, Haoyang Wen, and Ting Liu. 2019.
\newblock \href {https://doi.org/10.18653/v1/D19-1214} {A stack-propagation
  framework with token-level intent detection for spoken language
  understanding}.
\newblock In \emph{Proceedings of the 2019 Conference on Empirical Methods in
  Natural Language Processing and the 9th International Joint Conference on
  Natural Language Processing (EMNLP-IJCNLP)}, pages 2078--2087, Hong Kong,
  China. Association for Computational Linguistics.

\bibitem[{Rajpurkar et~al.(2016)Rajpurkar, Zhang, Lopyrev, and
  Liang}]{Rajpurkar2016SQuAD10}
Pranav Rajpurkar, Jian Zhang, Konstantin Lopyrev, and Percy Liang. 2016.
\newblock \href {https://doi.org/10.18653/v1/D16-1264} {{SQ}u{AD}: 100,000+
  questions for machine comprehension of text}.
\newblock pages 2383--2392.

\bibitem[{Robertson and Zaragoza(2009)}]{robertson2009bm25}
Stephen~E. Robertson and Hugo Zaragoza. 2009.
\newblock \href {https://doi.org/10.1561/1500000019} {The probabilistic
  relevance framework: {BM25} and beyond}.
\newblock \emph{Foundations and Trends in Information Retrieval},
  3(4):333--389.

\bibitem[{Romeo et~al.(2018)Romeo, Da~San~Martino, Barr{\'o}n-Cede{\~n}o, and
  Moschitti}]{romeo-etal-2018-flexible}
Salvatore Romeo, Giovanni Da~San~Martino, Alberto Barr{\'o}n-Cede{\~n}o, and
  Alessandro Moschitti. 2018.
\newblock \href {https://doi.org/10.18653/v1/P18-4023} {A flexible, efficient
  and accurate framework for community question answering pipelines}.
\newblock In \emph{Proceedings of {ACL} 2018, System Demonstrations}, pages
  134--139, Melbourne, Australia. Association for Computational Linguistics.

\bibitem[{R{\"u}ckl{\'e} and Gurevych(2017)}]{ruckle-gurevych-2017-end}
Andreas R{\"u}ckl{\'e} and Iryna Gurevych. 2017.
\newblock \href {https://www.aclweb.org/anthology/P17-4004} {End-to-end
  non-factoid question answering with an interactive visualization of neural
  attention weights}.
\newblock In \emph{Proceedings of {ACL} 2017, System Demonstrations}, pages
  19--24, Vancouver, Canada. Association for Computational Linguistics.

\bibitem[{R{\"u}ckl{\'e} et~al.(2019{\natexlab{a}})R{\"u}ckl{\'e}, Moosavi, and
  Gurevych}]{rueckle:AAAI:2019}
Andreas R{\"u}ckl{\'e}, Nafise~Sadat Moosavi, and Iryna Gurevych.
  2019{\natexlab{a}}.
\newblock \href {https://doi.org/10.1609/aaai.v33i01.33016932} {Coala: A neural
  coverage-based approach for long answer selection with small data.}
\newblock In \emph{Proceedings of the 33rd AAAI Conference on Artificial
  Intelligence (AAAI 2019)}, pages 6932--6939.

\bibitem[{R{\"u}ckl{\'e} et~al.(2019{\natexlab{b}})R{\"u}ckl{\'e}, Moosavi, and
  Gurevych}]{ruckle-etal-2019-neural}
Andreas R{\"u}ckl{\'e}, Nafise~Sadat Moosavi, and Iryna Gurevych.
  2019{\natexlab{b}}.
\newblock \href {https://doi.org/10.18653/v1/D19-1171} {Neural duplicate
  question detection without labeled training data}.
\newblock In \emph{Proceedings of the 2019 Conference on Empirical Methods in
  Natural Language Processing and the 9th International Joint Conference on
  Natural Language Processing (EMNLP-IJCNLP)}, pages 1607--1617, Hong Kong,
  China. Association for Computational Linguistics.

\bibitem[{R{\"u}ckl{\'e} et~al.(2020)R{\"u}ckl{\'e}, Pfeiffer, and
  Gurevych}]{ruckle-etal-2020-multicqa}
Andreas R{\"u}ckl{\'e}, Jonas Pfeiffer, and Iryna Gurevych. 2020.
\newblock \href {https://doi.org/10.18653/v1/2020.emnlp-main.194}
  {{M}ulti{CQA}: Zero-shot transfer of self-supervised text matching models on
  a massive scale}.
\newblock In \emph{Proceedings of the 2020 Conference on Empirical Methods in
  Natural Language Processing (EMNLP)}, pages 2471--2486, Online. Association
  for Computational Linguistics.

\bibitem[{Singh et~al.(2018)Singh, Radhakrishna, Both, Shekarpour, Lytra,
  Usbeck, Vyas, Khikmatullaev, Punjani, Lange, Vidal, Lehmann, and
  Auer}]{10.1145/3178876.3186023}
Kuldeep Singh, Arun~Sethupat Radhakrishna, Andreas Both, Saeedeh Shekarpour,
  Ioanna Lytra, Ricardo Usbeck, Akhilesh Vyas, Akmal Khikmatullaev, Dharmen
  Punjani, Christoph Lange, Maria~Esther Vidal, Jens Lehmann, and S\"{o}ren
  Auer. 2018.
\newblock \href {https://doi.org/10.1145/3178876.3186023} {\emph{Why Reinvent
  the Wheel: Let's Build Question Answering Systems Together}}, page
  1247–1256. International World Wide Web Conferences Steering Committee,
  Republic and Canton of Geneva, CHE.

\bibitem[{Sorokin and Gurevych(2018)}]{sorokin-gurevych-2018-interactive}
Daniil Sorokin and Iryna Gurevych. 2018.
\newblock \href {https://doi.org/10.18653/v1/D18-2020} {Interactive
  instance-based evaluation of knowledge base question answering}.
\newblock In \emph{Proceedings of the 2018 Conference on Empirical Methods in
  Natural Language Processing: System Demonstrations}, pages 114--119,
  Brussels, Belgium. Association for Computational Linguistics.

\bibitem[{Sun et~al.(2019)Sun, Bedrax{-}Weiss, and
  Cohen}]{sun-etal-2019-pullnet}
Haitian Sun, Tania Bedrax{-}Weiss, and William~W. Cohen. 2019.
\newblock \href {https://doi.org/10.18653/v1/D19-1242} {Pull{N}et: Open domain
  question answering with iterative retrieval on knowledge bases and text}.
\newblock In \emph{Proceedings of the 2019 Conference on Empirical Methods in
  Natural Language Processing and the 9th International Joint Conference on
  Natural Language Processing, {EMNLP-IJCNLP} 2019, Hong Kong, China, November
  3-7, 2019}, pages 2380--2390. Association for Computational Linguistics.

\bibitem[{Sun et~al.(2018)Sun, Dhingra, Zaheer, Mazaitis, Salakhutdinov, and
  Cohen}]{sun2018}
Haitian Sun, Bhuwan Dhingra, Manzil Zaheer, Kathryn Mazaitis, Ruslan
  Salakhutdinov, and William~W. Cohen. 2018.
\newblock \href {https://doi.org/10.18653/v1/d18-1455} {Open domain question
  answering using early fusion of knowledge bases and text}.
\newblock In \emph{Proceedings of the 2018 Conference on Empirical Methods in
  Natural Language Processing, Brussels, Belgium, October 31 - November 4,
  2018}, pages 4231--4242. Association for Computational Linguistics.

\bibitem[{Talmor and Berant(2019)}]{talmor-berant-2019-multiqa}
Alon Talmor and Jonathan Berant. 2019.
\newblock \href {https://doi.org/10.18653/v1/P19-1485} {{M}ulti{QA}: An
  empirical investigation of generalization and transfer in reading
  comprehension}.
\newblock In \emph{Proceedings of the 57th Annual Meeting of the Association
  for Computational Linguistics}, pages 4911--4921, Florence, Italy.
  Association for Computational Linguistics.

\bibitem[{Tay et~al.(2017)Tay, Phan, Luu, and Hui}]{Tay2017}
Yi~Tay, Minh~C. Phan, Anh~Tuan Luu, and Siu~Cheung Hui. 2017.
\newblock \href {https://doi.org/10.1145/3077136.3080790} {Learning to rank
  question answer pairs with holographic dual {LSTM} architecture}.
\newblock In \emph{Proceedings of the 40th International {ACM} {SIGIR}
  Conference on Research and Development in Information Retrieval, Shinjuku,
  Tokyo, Japan, August 7-11, 2017}, pages 695--704. {ACM}.

\bibitem[{Usbeck et~al.(2017)Usbeck, Ngomo, Haarmann, Krithara, R{\"{o}}der,
  and Napolitano}]{10.1007/978-3-319-69146-6_6}
Ricardo Usbeck, Axel{-}Cyrille~Ngonga Ngomo, Bastian Haarmann, Anastasia
  Krithara, Michael R{\"{o}}der, and Giulio Napolitano. 2017.
\newblock \href {https://doi.org/10.1007/978-3-319-69146-6\_6} {7th open
  challenge on question answering over linked data {(QALD-7)}}.
\newblock In \emph{Semantic Web Challenges - 4th SemWebEval Challenge at {ESWC}
  2017, Portoroz, Slovenia, May 28 - June 1, 2017, Revised Selected Papers},
  volume 769 of \emph{Communications in Computer and Information Science},
  pages 59--69. Springer.

\bibitem[{Vaswani et~al.(2017)Vaswani, Shazeer, Parmar, Uszkoreit, Jones,
  Gomez, Kaiser, and Polosukhin}]{vaswani17attention}
Ashish Vaswani, Noam Shazeer, Niki Parmar, Jakob Uszkoreit, Llion Jones,
  Aidan~N. Gomez, undefinedukasz Kaiser, and Illia Polosukhin. 2017.
\newblock Attention is all you need.
\newblock In \emph{Proceedings of the 31st International Conference on Neural
  Information Processing Systems}, NIPS'17, page 6000–6010, Red Hook, NY,
  USA. Curran Associates Inc.

\bibitem[{Xu et~al.(2016)Xu, Feng, Huang, and Zhao}]{xu-etal-2016-hybrid}
Kun Xu, Yansong Feng, Songfang Huang, and Dongyan Zhao. 2016.
\newblock \href {https://www.aclweb.org/anthology/C16-1226/} {Hybrid question
  answering over knowledge base and free text}.
\newblock In \emph{{COLING} 2016, 26th International Conference on
  Computational Linguistics, Proceedings of the Conference: Technical Papers,
  December 11-16, 2016, Osaka, Japan}, pages 2397--2407. {ACL}.

\bibitem[{Yang et~al.(2019)Yang, Xie, Lin, Li, Tan, Xiong, Li, and
  Lin}]{yang-etal-2019-end-end-open}
Wei Yang, Yuqing Xie, Aileen Lin, Xingyu Li, Luchen Tan, Kun Xiong, Ming Li,
  and Jimmy Lin. 2019.
\newblock \href {https://doi.org/10.18653/v1/N19-4013} {End-to-end open-domain
  question answering with {BERT}serini}.
\newblock In \emph{Proceedings of the 2019 Conference of the North {A}merican
  Chapter of the Association for Computational Linguistics (Demonstrations)},
  pages 72--77, Minneapolis, Minnesota. Association for Computational
  Linguistics.

\bibitem[{Yang et~al.(2020)Yang, Cer, Ahmad, Guo, Law, Constant, Abrego, Yuan,
  Tar, Sung, Strope, and Kurzweil}]{Yang2020MultilingualUS}
Yinfei Yang, Daniel Cer, Amin Ahmad, Mandy Guo, Jax Law, Noah Constant,
  Gustavo~Hernandez Abrego, Steve Yuan, Chris Tar, Yun-hsuan Sung, Brian
  Strope, and Ray Kurzweil. 2020.
\newblock \href {https://doi.org/10.18653/v1/2020.acl-demos.12} {Multilingual
  universal sentence encoder for semantic retrieval}.
\newblock pages 87--94.

\bibitem[{Yang et~al.(2018)Yang, Qi, Zhang, Bengio, Cohen, Salakhutdinov, and
  Manning}]{Yang2018HotpotQAAD}
Zhilin Yang, Peng Qi, Saizheng Zhang, Yoshua Bengio, William Cohen, Ruslan
  Salakhutdinov, and Christopher~D. Manning. 2018.
\newblock \href {https://doi.org/10.18653/v1/D18-1259} {{H}otpot{QA}: A dataset
  for diverse, explainable multi-hop question answering}.
\newblock pages 2369--2380.

\bibitem[{Yih et~al.(2016)Yih, Richardson, Meek, Chang, and
  Suh}]{yih-etal-2016-value}
Wen-tau Yih, Matthew Richardson, Chris Meek, Ming-Wei Chang, and Jina Suh.
  2016.
\newblock \href {https://doi.org/10.18653/v1/P16-2033} {The value of semantic
  parse labeling for knowledge base question answering}.
\newblock In \emph{Proceedings of the 54th Annual Meeting of the Association
  for Computational Linguistics (Volume 2: Short Papers)}, pages 201--206,
  Berlin, Germany. Association for Computational Linguistics.

\bibitem[{Zhao et~al.(2018)Zhao, Zhang, Duan, Chen, Zhang, and
  Tang}]{zhao2018novel}
Shu Zhao, Dong Zhang, Zhen Duan, Jie Chen, Yan{-}Ping Zhang, and Jie Tang.
  2018.
\newblock \href {https://doi.org/10.1007/s11192-018-2726-6} {A novel
  classification method for paper-reviewer recommendation}.
\newblock \emph{Scientometrics}, 115(3):1293--1313.

\end{thebibliography}
\bibliographystyle{acl_natbib}

\clearpage
\appendix

\section{Appendix}
\subsection{Inference Time Comparison between Models}
\label{section:inference}
Our system should be computationally efficient, i.e., its latency should be low even with a large quantity of QA agents.
To estimate the computational efficiency of our approaches,
we measure how many queries per second (qps) our model can complete with 10 and 200 agents.
TWEAC is tested with both large and base-sized transformers.
We index 1000 examples per agent for the similarity-based methods.
BM25 is implemented using Elasticsearch\footnote{\url{https://www.elastic.co}} with the default configuration and we use Faiss  with a flat index for GPU-based kNN lookup with USE-QA.
TWEAC and USE-QA process each question separately without batching to simulate single questions coming in from a user.
GPU computation is done on an NVIDIA Titan-RTX GPU.

\begin{table}[h]
    \centering
    \footnotesize
\begin{tabular}{rlr}
\toprule
\bf QA Agents &          \bf Name &  \bf QPS \\
\midrule
      10 &  TWEAC{\scriptsize{R-l}} &  49.44$\pm$\phantom{0}0.15  \\
      10 & TWEAC{\scriptsize{A-l}} & 39.44$\pm$\phantom{0}0.15 \\
      10 &    TWEAC{\scriptsize{A-b}} &  76.97$\pm$\phantom{0}1.15  \\
      10 &         USE-QA &  43.60$\pm$\phantom{0}5.01  \\
      10 &           BM25 &  232.29$\pm$60.19  \\
     \midrule
     200 &  TWEAC{\scriptsize{R-l}} &  46.41$\pm$\phantom{0}2.00  \\
     200 & TWEAC{\scriptsize{A-l}} & 36.39$\pm$\phantom{0}0.30 \\
     200 &    TWEAC{\scriptsize{A-b}} &  70.45$\pm$\phantom{0}0.76  \\
     200 &         USE-QA &  48.10$\pm$\phantom{0}1.67 \\
     200 &           BM25 &  221.19$\pm$34.52  \\
\bottomrule
\end{tabular}
    \caption{Queries per second (qps) of the different methods with 10 and 200 agents. The results are the mean and standard deviation over 5000 and 100.000 iterations for the 10 and 200 agents respectively.}
    \label{tab:qps}
\end{table}
Table~\ref{tab:qps} shows the results of the measurement.
We observe that Elasticsearch is by far the fastest.
USE-QA is similar in speed ($\pm$ 10 qps) to the large TWEAC models.

The choice of transformer architecture has a large impact on TWEAC.
TWEAC{\scriptsize{A-b}} is roughly twice as fast as TWEAC{\scriptsize{A-l}} which also slower than TWEAC{\scriptsize{R-l}} as well.
We note that there is only a slight increase in the inference time when using TWEAC with 10 and 200 agents.

In conclusion, the model choice presents a trade-off between speed and accuracy:
While Elasticsearch with BM25 is considerably faster and requires no GPU, it is also vastly less accurate than the other methods as shown by our experiments.
TWEAC{\scriptsize{A-b}} is 50\% faster than TWEAC{\scriptsize{R-l}} but the latter has a higher performance in all experiments.

\subsection{Figure Values}
\label{section:tablesval}

In this section, we present the results to Figure 3, 4 and 5 in table format along with MRR, which is not included in the figures.
\begin{table}[htp]
\footnotesize
\begin{tabular}{rlrr}
\toprule
 \bf Examples & \bf Name &  \bf Accuracy &  \bf MRR \\
\midrule
      64 &  TWEAC{\scriptsize{R-l}} &    0.8572 & 0.9153 \\
      64 &  TWEAC{\scriptsize{A-l}} &    0.8449 & 0.9028 \\
      64 &  TWEAC{\scriptsize{A-b}} &    0.8271 & 0.8938 \\
      64 &                   USE-QA &    0.7143 & 0.8314 \\
      64 &                     BM25 &    0.6057 & 0.7457 \\
      \midrule
     128 &  TWEAC{\scriptsize{R-l}} &    0.8934 & 0.9380 \\
     128 &  TWEAC{\scriptsize{A-l}} &    0.8646 & 0.9174 \\
     128 &  TWEAC{\scriptsize{A-b}} &    0.8566 & 0.9089 \\
     128 &                   USE-QA &    0.7562 & 0.8582 \\
     128 &                     BM25 &    0.6496 & 0.7762 \\
      \midrule
     256 &  TWEAC{\scriptsize{R-l}} &    0.9158 & 0.9519 \\
     256 &  TWEAC{\scriptsize{A-l}} &    0.8980 & 0.9392 \\
     256 &  TWEAC{\scriptsize{A-b}} &    0.8600 & 0.9074 \\
     256 &                   USE-QA &    0.8006 & 0.8826 \\
     256 &                     BM25 &    0.6939 & 0.8093 \\
      \midrule
     512 &  TWEAC{\scriptsize{R-l}} &    0.9303 & 0.9606 \\
     512 &  TWEAC{\scriptsize{A-l}} &    0.9139 & 0.9497 \\
     512 &  TWEAC{\scriptsize{A-b}} &    0.8621 & 0.9173 \\
     512 &                   USE-QA &    0.8189 & 0.8946 \\
     512 &                     BM25 &    0.7078 & 0.8226 \\
      \midrule
    1024 &  TWEAC{\scriptsize{R-l}} &    0.9371 & 0.9638 \\
    1024 &  TWEAC{\scriptsize{A-l}} &    0.9275 & 0.9577 \\
    1024 &  TWEAC{\scriptsize{A-b}} &    0.9096 & 0.9462 \\
    1024 &                   USE-QA &    0.8441 & 0.9092 \\
    1024 &                     BM25 &    0.7385 & 0.8407 \\
      \midrule
    2048 &  TWEAC{\scriptsize{R-l}} &    0.9428 & 0.9677 \\
    2048 &  TWEAC{\scriptsize{A-l}} &    0.9266 & 0.9581 \\
    2048 &  TWEAC{\scriptsize{A-b}} &    0.8998 & 0.9409 \\
    2048 &                   USE-QA &    0.8520 & 0.9149 \\
    2048 &                     BM25 &    0.7609 & 0.8540 \\
      \midrule
    4096 &  TWEAC{\scriptsize{R-l}} &    0.9508 & 0.9723 \\
    4096 &  TWEAC{\scriptsize{A-l}} &    0.9324 & 0.9606 \\
    4096 &  TWEAC{\scriptsize{A-b}} &    0.9031 & 0.9415 \\
    4096 &                   USE-QA &    0.8441 & 0.9092 \\
    4096 &                     BM25 &    0.7385 & 0.8407 \\

\bottomrule
\end{tabular}
    \caption{Table to Figure 3}
    \label{tab:sample}
\end{table}

\begin{table}[htp]
    \centering
\footnotesize
    \subfloat[reddit]{
\begin{tabular}{rlrr}
\toprule
\bf QA Agents &   \bf Name &  \bf Accuracy & \bf MRR \\
\midrule
      10 &  TWEAC{\scriptsize{R-l}} &    0.8346 & 0.8902 \\
      10 &   TWEAC{\scriptsize{A-l}} &    0.7746 & 0.8497 \\
      10 &    TWEAC{\scriptsize{A-b}} &    0.7662 & 0.8436 \\
      10 &         USE-QA &    0.7654 & 0.8500 \\
      10 &           BM25 &    0.6664 & 0.7682 \\
      \midrule
      50 &  TWEAC{\scriptsize{R-l}} &    0.6898 & 0.7638 \\
      50 &   TWEAC{\scriptsize{A-l}} &    0.6042 & 0.6967 \\
      50 &    TWEAC{\scriptsize{A-b}} &    0.5980 & 0.6874 \\
      50 &         USE-QA &    0.6232 & 0.7141 \\
      50 &           BM25 &    0.4855 & 0.5824 \\
      \midrule
     100 &  TWEAC{\scriptsize{R-l}} &    0.6029 & 0.6872 \\
     100 &   TWEAC{\scriptsize{A-l}} &    0.5146 & 0.6186 \\
     100 &    TWEAC{\scriptsize{A-b}} &    0.5145 & 0.6049 \\
     100 &         USE-QA &    0.5469 & 0.6394 \\
     100 &           BM25 &    0.4186 & 0.5080 \\
     \midrule
     200 &  TWEAC{\scriptsize{R-l}} &    0.5226 & 0.6173 \\
     200 &   TWEAC{\scriptsize{A-l}} &    0.4463 & 0.5513 \\
     200 &    TWEAC{\scriptsize{A-b}} &    0.4428 & 0.5385 \\
     200 &         USE-QA &    0.4813 & 0.5734 \\
     200 &           BM25 &    0.3604 & 0.4437 \\
\bottomrule
\end{tabular}}\\
\subfloat[StackExchange]{
\begin{tabular}{rlrr}
\toprule
\bf QA Agents &   \bf Name &  \bf Accuracy & \bf MRR \\
\midrule
      10 &  TWEAC{\scriptsize{R-l}} &    0.8648 & 0.9135 \\
      10 &   TWEAC{\scriptsize{A-l}} &    0.8582 & 0.9072 \\
      10 &    TWEAC{\scriptsize{A-b}} &    0.8377 & 0.8964 \\
      10 &         USE-QA &    0.8812 & 0.9275 \\
      10 &           BM25 &    0.7572 & 0.8453 \\
      \midrule
      50 &  TWEAC{\scriptsize{R-l}} &    0.7339 & 0.8093 \\
      50 &   TWEAC{\scriptsize{A-l}} &    0.6947 & 0.7768 \\
      50 &    TWEAC{\scriptsize{A-b}} &    0.6655 & 0.7543 \\
      50 &         USE-QA &    0.7046 & 0.7952 \\
      50 &           BM25 &    0.5707 & 0.6827 \\
      \midrule
     100 &  TWEAC{\scriptsize{R-l}} &    0.6830 & 0.7708 \\
     100 &   TWEAC{\scriptsize{A-l}} &    0.6238 & 0.7239 \\
     100 &    TWEAC{\scriptsize{A-b}} &    0.5914 & 0.6966 \\
     100 &         USE-QA &    0.6321 & 0.7343 \\
     100 &           BM25 &    0.5027 & 0.6128 \\
     \midrule
     171 &  TWEAC{\scriptsize{R-l}} &    0.6285 & 0.7250 \\
     171 &   TWEAC{\scriptsize{A-l}} &    0.5604 & 0.6705 \\
     171 &    TWEAC{\scriptsize{A-b}} &    0.5258 & 0.6359 \\
     171 &         USE-QA &    0.5636 & 0.6733 \\
     171 &           BM25 &    0.4476 & 0.5544 \\
\bottomrule
\end{tabular}}

    \caption{Table to Figure 4.}
    \label{tab:setup2comp}
\end{table}

\begin{table}[htp]
\footnotesize
    \centering
\subfloat[reddit]{
\begin{tabular}{rlrr}
\toprule
\bf QA Agents &   \bf Name &  \bf Accuracy & \bf MRR \\
\midrule
      50 &  full training &    0.5980 & 0.6874 \\
      50 &    no sampling (10)&    0.5028 & 0.5664 \\
      50 &  half-and-half (10)&    0.5556 & 0.6505 \\
\midrule
     100 &  full training &    0.5145 & 0.6049 \\
     100 &    no sampling (10) &    0.3971 & 0.4542 \\
     100 &    no sampling (50) &    0.4363 & 0.4954 \\
     100 &  half-and-half (10)&    0.4406 & 0.5421 \\
     100 &  half-and-half (50)&    0.4579 & 0.5630 \\
\bottomrule
\end{tabular}}

\subfloat[StackExchange]{
\begin{tabular}{rlrr}
\toprule
\bf QA Agents &   \bf Name &  \bf Accuracy & \bf MRR \\
\midrule
      50 &  full training &    0.6655 & 0.7543 \\
      50 &    no sampling &    0.6447 & 0.7296 \\
      50 &  half-and-half &    0.6511 & 0.7458 \\
\midrule
     100 &  full training &    0.5914 & 0.6966 \\
     100 &    no sampling (10)&    0.5541 & 0.6329 \\
     100 &    no sampling (50)&    0.5900 & 0.6846 \\
     100 &  half-and-half (10) &    0.5557 & 0.6738 \\
     100 &  half-and-half (50)&    0.5696 & 0.6834 \\
\bottomrule
\end{tabular}}
    \caption{Table to Figure 5 with values for 50 and 100 agents. The starting number of agents for the iterative extension models is indicated in parenthesis.}
    \label{tab:it_ext}
\end{table}

\subsection{Misclassification Analysis}
\label{section:tableanalysis}
In this section, we present the tables to \S7.
\begin{table}[htp]
\footnotesize
    \centering
    \begin{tabular}{lr}
    \toprule
    \multicolumn{2}{l}{\bf reddit}\\
      Total unique mistakes   & 4999  \\
        Total misclassification errors     & 32523 \\
        0.540\% of mistakes cause 10\% of errors &\\
        8.801\% of mistakes cause 50\% of errors & \\
        51.45\% of mistakes cause 90\% of errors  &\\
    \midrule
    \multicolumn{2}{l}{\bf StackExchanges}\\
      Total unique mistakes   & 9120  \\
        Total misclassification errors     & 48882 \\
        0.394\% of mistakes cause 10\% of errors &\\
        9.736\% of mistakes cause 50\% of errors &\\
        59.21\% of mistakes cause 90\% of errors &\\
    \bottomrule
    \end{tabular}
    \caption{Statistics to the misclassification errors and the unique mistakes between agents in \S7}
    \label{tab:errorstat}
\end{table}

\begin{table}[htp]
    \centering
\footnotesize
    \subfloat[reddit]{
    \begin{tabular}{lr}
    \toprule
    \bf Mistake & \bf \permil \, of total errors  \\
    \midrule
politics $\leftrightarrow$ Conservative & 5.50 \\
recipes $\leftrightarrow$ food & 5.09 \\
Autos $\leftrightarrow$ cars & 4.15 \\
unitedkingdom $\leftrightarrow$ ukpolitics & 4.09 \\
apple $\leftrightarrow$ mac & 3.72 \\
beer $\leftrightarrow$ Coffee & 3.21 \\
Anarchism $\leftrightarrow$ socialism & 3.15 \\
dating $\leftrightarrow$ seduction & 3.07 \\
snowboarding $\leftrightarrow$ skiing & 2.97 \\
Games $\leftrightarrow$ pcgaming & 2.95 \\
techsupport $\leftrightarrow$ computers & 2.72 \\
bicycling $\leftrightarrow$ motorcycles & 2.66 \\
soccer $\leftrightarrow$ sports & 2.66 \\
atheism $\leftrightarrow$ Christianity & 2.58 \\
apple $\leftrightarrow$ technology & 2.52 \\
Games $\leftrightarrow$ videogames & 2.50 \\
Buddhism $\leftrightarrow$ zen & 2.23 \\
horror $\leftrightarrow$ movies & 2.17 \\
Marijuana $\leftrightarrow$ weed & 2.17 \\
techsupport $\leftrightarrow$ windows & 2.17 \\
Total Sum & 62.27 \\
\bottomrule
    \end{tabular}}\\
    \subfloat[StackExchange]{
    \begin{tabular}{lr}
    \toprule
    \bf Mistake & \bf \permil \, of total errors \\
    \midrule
english $\leftrightarrow$ ell & 8.49 \\
cstheory $\leftrightarrow$ cs & 5.50 \\
space $\leftrightarrow$ astronomy & 5.17 \\
hermeneutics $\leftrightarrow$ christianity & 4.64 \\
korean $\leftrightarrow$ beer & 4.52 \\
scifi $\leftrightarrow$ movies & 4.37 \\
windowsphone $\leftrightarrow$ android & 4.00 \\
linguistics $\leftrightarrow$ conlang & 3.90 \\
elementaryos $\leftrightarrow$ askubuntu & 3.87 \\
stats $\leftrightarrow$ datascience & 3.87 \\
superuser $\leftrightarrow$ serverfault & 3.63 \\
monero $\leftrightarrow$ bitcoin & 3.57 \\
homebrew $\leftrightarrow$ beer & 3.26 \\
matheducators $\leftrightarrow$ math & 3.26 \\
security $\leftrightarrow$ crypto & 3.26 \\
unix $\leftrightarrow$ askubuntu & 3.26 \\
skeptics $\leftrightarrow$ health & 3.14 \\
vi $\leftrightarrow$ emacs & 2.98 \\
movies $\leftrightarrow$ literature & 2.92 \\
physics $\leftrightarrow$ astronomy & 2.89 \\
Total Sum & 80.50 \\
\bottomrule
    \end{tabular}}

    \caption{Twenty largest mistakes in \S7}
    \label{tab:top20mistakes}
\end{table}

\end{document}